\documentclass[10pt, a4paper]{article}

\usepackage{lrec-coling2024} 
\usepackage{tabularx}
\usepackage{graphicx}
\usepackage{multirow}
\usepackage{booktabs} 
\usepackage{subfig}
\newcommand*{\myfont}{\fontfamily{Times New Roman}\selectfont}

\title{LFED: A Literary Fiction Evaluation Dataset for Large Language Models}

\name{Linhao Yu$^1$, Qun Liu$^2$, Deyi Xiong$^{1\ast}$ \thanks{*Corresponding author}} 

\address{
          $^1$College of Intelligence and Computing, Tianjin University, Tianjin, China \\ 
          $^2$Huawei Noah's Ark Lab \\
         \{linhaoyu, dyxiong\}@tju.edu.cn, qun.liu@huawei.com\\}

\abstract{
The rapid evolution of large language models (LLMs) has ushered in the need for comprehensive assessments of their performance across various dimensions. In this paper, we propose LFED, a \textbf{L}iterary \textbf{F}iction \textbf{E}valuation \textbf{D}ataset, which aims to evaluate the capability of LLMs on the long fiction comprehension and reasoning. We collect 95 literary fictions that are either originally written in Chinese or translated into Chinese, covering a wide range of topics across several centuries. We define a question taxonomy with 8 question categories to guide the creation of 1,304 questions.  Additionally, we conduct an in-depth analysis to ascertain how specific attributes of literary fictions (e.g., novel types, character numbers, the year of publication) impact LLM performance in evaluations.  Through a series of experiments with various state-of-the-art LLMs, we demonstrate that these models face considerable challenges in effectively addressing questions related to literary fictions, with ChatGPT reaching only 57.08\% under the zero-shot setting. The dataset will be publicly available at \url{https://github.com/tjunlp-lab/LFED.git}. 
 \\ \newline  }

\begin{document}

\maketitleabstract

\section{Introduction}

Numerous datasets have been developed to facilitate machine reading comprehension tasks, e.g., MCTest \citep{DBLP:conf/emnlp/RichardsonBR13}, MCScript \citep{DBLP:conf/lrec/0002MRTP18}, RACE \citep{DBLP:conf/emnlp/LaiXLYH17}, CoQA \citep{DBLP:journals/tacl/ReddyCM19}, WYWEB \citep{DBLP:conf/acl/ZhouCZZ23}, to name a few. However, as large language models (LLMs) have made remarkable progress recently, these passage-based datasets are no longer capable of evaluating such large models. More challenging datasets with long documents that go beyond the context windows of LLMs (even for the 100K-token context window of Anthropic Claude\footnote{https://www.anthropic.com/index/100k-context-windows}), complicated reasoning (e.g., character relationship reasoning, counterfactual reasoning), skills of connoisseurship, etc., are much desirable for evaluating highly capable LLMs. 

To bridge this gap, we curate LFED, a Literary Fiction Evaluation Dataset for large language models. LFED is a comprehensive dataset derived from a diverse collection of literary fictions that are either originally written in Chinese or translated into Chinese. It encompasses 8 distinct question types, which focus on the core aspects of the fictions, such as content, character relationships, storyline, writing techniques, and thematic values. In order to automate and standardize the evaluation of LLMs on LFED, we construct multiple-choice questions under each question type, providing carefully-prepared multiple answer choices for each question. The construction of the dataset is via crowdsourcing, with rigorous quality control. Ultimately, we have curated a total of 1,304 questions derived from 95 fictions. This dataset can serve as a comprehensive and challenging evaluation benchmark for assessing the fact understanding, logical reasoning, contextual comprehension, common-sense reasoning, and value judgment capabilities of large language models.

Our main contributions are summarized as follows.
\begin{enumerate}
	\item We propose LFED, which, to the best of our knowledge, is the first Chinese dataset curated for evaluating LLMs on long literary fictions. 
	\item We define a question taxonomy according to the nature of literary fictions, which exhibits a wide coverage on the skills necessary for reading and understanding these fictions.
	\item We have evaluated a number of LLMs on the curated dataset under the zero- and few-shot setting. Evaluation results demonstrate that long literary fiction comprehension is very challenging for LLMs, with ChatGPT achieving an accuracy of 57.08\% under the zero-shot setting.
\end{enumerate}
\begin{figure*}[!ht]
	\centering
	\includegraphics[width = \linewidth]{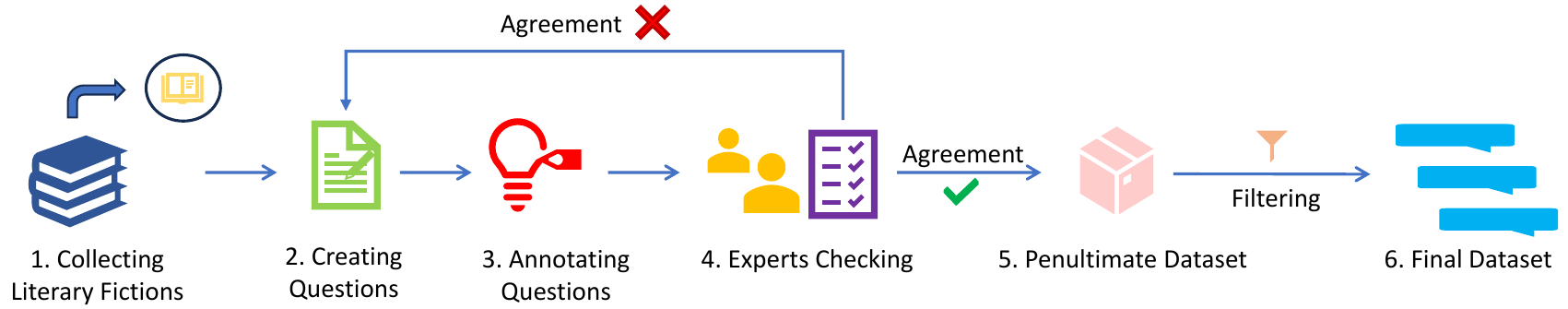}
	\caption{The overall pipeline for collecting questions in LFED.}
	\label{fig: workflow}
\end{figure*}

\section{Related Work}
We review existing machine reading comprehension (MRC) and question answering (QA) datasets within the scope and page constraint of this paper, highlighting representative Chinese datasets in different categories \citep{DBLP:journals/corr/abs-2310-19736}.
\textbf{Multiple-Choice QA Datasets} Multiple-choice questions are a specific question format that provides answer choices for each question. Numerous existing multiple-choice QA datasets are sourced from school examinations. For instance, RACE \citep{DBLP:conf/emnlp/LaiXLYH17} encompasses a vast collection of over 28,000 essays and nearly 100,000 questions, extracting from both general and specific subjects covered in Chinese middle and high school English exams. NCR \citep{DBLP:conf/nips/XuLYZLW21}, on the other hand, comprises remarkably long modern and classical Chinese essays on various topics derived from high school Chinese language courses. It is tailored to evaluate the language proficiencies of native speakers. MCTest \citep{DBLP:conf/emnlp/RichardsonBR13} presents single-choice reading comprehension questions based on fictional stories.  Additionally, recent efforts have been dedicated to curating datasets in the multiple-choice QA form for evaluating LLMs from different perspectives \citep{DBLP:journals/corr/abs-2310-19736}, such as CBBQ \citep{DBLP:journals/corr/abs-2306-16244},  covering stereotypes and societal biases in 14 social dimensions related to Chinese culture and values, RoleEval \citep{DBLP:journals/corr/abs-2312-16132}, a bilingual benchmark designed to assess the memorization, utilization, and reasoning capabilities of role knowledge, etc. M3KE \citep{DBLP:journals/corr/abs-2305-10263} collects 20,477 questions from 71 tasks, covering all major levels of the Chinese education system, from primary school to university, and a wide range of subjects including humanities, history, politics, law, education, psychology, science, technology, art and religion. LHMKE\citep{liu2024lhmke} encompasses 10,465 questions across 75 tasks covering 30 subjects, ranging from primary school to professional certification exams. Notably, LHMKE includes both objective and subjective questions, offering a more holistic evaluation of the knowledge level of LLMs. 

\textbf{Extractive MRC Datasets} There has been a significant surge in the development of various extractive MRC datasets. One prominent example is SQuAD \citep{DBLP:conf/emnlp/RajpurkarZLL16}, which comprises questions generated by crowdsourced workers based on a collection of Wikipedia passages. Each question is designed to elicit an answer that corresponds to a specific text or span within the associated reading passage. Another dataset is BiPaR \citep{DBLP:conf/emnlp/JingXZ19}, which is a manually annotated bilingual parallel novel machine reading comprehension dataset. It facilitates monolingual, multilingual, and interlingual reading comprehension tasks specifically focused on novels. CMRC2018 \citep{DBLP:conf/emnlp/CuiLCXCMWH19}, on the other hand, is an extractive dataset designed for Chinese machine reading comprehension. It contains a substantial collection of 20,000 real-world questions derived from Wikipedia sources. Furthermore, CJRC \citep{DBLP:conf/cncl/DuanWWMCWWLHHWL19} is a dataset specifically created for Chinese judicial reading comprehension. The documents in this dataset are sourced from judicial documents, and the questions are annotated by legal experts, providing a valuable resource for exploring legal domain comprehension tasks.

\textbf{Generative MRC Datasets} The most authentic approach for human question answering involves generating answers independently, without being constrained to selecting predetermined options or extracting fragments from given documents as answers. This format enables the exploration of various question types. Notably, MS MARCO \citep{DBLP:conf/nips/NguyenRSGTMD16} is designed as a generative dataset that emphasizes deep learning in the search domain. In the case of NarrativeQA \citep{DBLP:journals/tacl/KociskySBDHMG18}, questions and answers are crafted by crowdsourcing workers based on book summaries. This format necessitates models to comprehend the underlying narrative in order to provide accurate answers. Additionally, DRCD \citep{DBLP:journals/corr/abs-1806-00920} serves as a standard Chinese machine reading comprehension dataset. It consists of 10,014 paragraphs sourced from 2,108 Wikipedia articles, accompanied by over 30,000 questions generated by annotators.

Our LFED is unique in its utilization of long literary fictions as the data source, deviating from passage-based QA datasets. Furthermore, LFED offers a comprehensive assessment of LLMs capabilities in fact understanding, logical reasoning, context comprehension, common sense reasoning, and value judgment across eight distinct question categories.

\section{Dataset Creation}
Figure \ref{fig: workflow} shows the overall dataset annotation pipeline. We design a very rigorous annotation process to ensure the quality of the dataset at each step, from the source of the dataset to the annotation and review of the dataset. We also design a question taxonomy to guide annotation.
\subsection{Data Source}
A wide variety of novels are selected according their complex narratives, character development, profound themes, and rich linguistic expressions. These aspects make novels suitable for evaluating Large Language Models (LLMs) in various capacities, including fact understanding and logical reasoning. Unlike academic articles, which are structured and precise, or news reports, which are concise and direct, novels provide a deeper, more nuanced content that challenges LLMs to understand underlying themes and cultural nuances.

We crawl a list of the top 200 literary novels' name according to the recommendations on Douban\footnote{\url{https://book.douban.com/}}, a Chinse community site with reviews of books and movies. We only select literary novels according to the reviews of readers published in Douban. We do NOT use any electronic versions of these fictions. All our hired crowdsourced workers read these fictions either with copyright or with bought hardcopies. Crowdsourced workers creating these questions give informed consent for the use of their contributions in LLM evaluation. Subsequently, each literary fiction in the list is manually checked to see if it satisfies with specific requirements. Only those that pass this manual selection are kept. The specific requirements are as follows:

\begin{itemize}
	\item Choosing classic novels: Classic novels usually have literary, historical and cultural values, which are widely recognized and read.
	\item Considering the genre of fictions: Such a consideration aims to diversify the selected literary fictions in terms of genres. 
	\item Scrutinizing Fiction content:The selected novel should be in line with human values. we eliminate novels that do not conform to contemporary values in the screening process, even though no such novels are ultimately selected. But in doing so, our dataset can prevent possible bias and ethical problems that current LLMs attempt to avoid too.
	\item Taking the popularity and influence of a fiction into account: The popularity of a fiction would make it easy for us to find crowdsourced workers to create questions and answers for it. 
\end{itemize}

\begin{figure}[!ht]
	\centering
	\includegraphics[width = \linewidth]{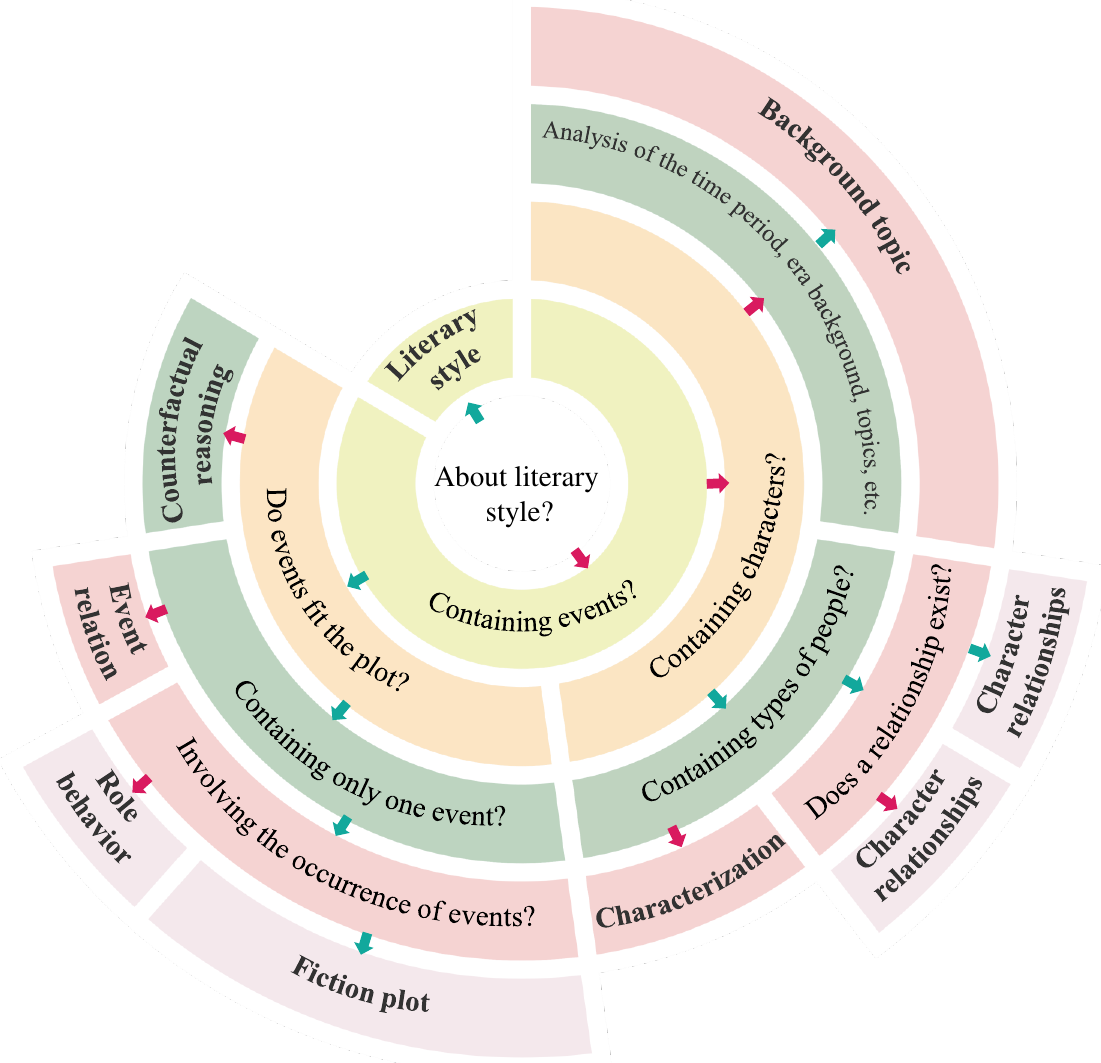}
	\caption{\label{fig: decision tree}Decision-tree-style Illustration of the question taxonomy. Green arrows denote yes, while red arrows indicate no.}
\end{figure}

\subsection{Question Taxonomy }\label{subsec: Question Taxonomy}
We develop a question taxonomy to guide the creation of questions according to the nature and characteristics of literary fictions. The taxonomy covers 8 question categories, illustrated in Figure \ref{fig: decision tree}, which are character relationships, characterization, literary style, role behavior, event relations, fiction plot, background topic, and counterfactual reasoning. We provide the descriptions and examples of the 8 question categories in Table \ref{tab: question-classify}. 

The design of the question taxonomy follows a systematic investigation of characters and events featured in literary fictions, which aims at the diversity and coverage of curated questions in the dataset.

\begin{table*}[!ht]
	\centering
	\begin{tabularx}{\textwidth} { 
			>{\raggedright\arraybackslash\hsize=.6\hsize\linewidth=\hsize}X
			>{\raggedright\arraybackslash\hsize=1\hsize\linewidth=\hsize}X
			>{\raggedright\arraybackslash\hsize=1.4\hsize\linewidth=\hsize}X
		}
		\toprule
		\textbf{Q. Category} & \textbf{Description}  & \textbf{Example}\\
		\midrule
		Character relationships & Relationships between two characters, such as master and apprentice, lovers, and so on. & Regarding the fiction \textit{\textquotedblleft The Return of the Condor Heroes\textquotedblright \ }, who is Yang Guo's favorite master? 
		\linebreak \myfont{A. Little Dragon girl B. Huang Rong C. Guo Jing D. Master Jin Lun} \\
		\midrule
		Characterization & The emotional transformation and personality change of a character in the story. & Regarding the novel \textit{\textquotedblleft Pride and Prejudice\textquotedblright \ }, what are the character traits of Mr. Darcy? \linebreak \myfont{A. He is arrogant B. He is ruthless C. He is cold D. He is kind} \\
		\midrule
		Literary style & The literary style , e.g., expository, narrative.& Regarding the fiction \textit{\textquotedblleft White Night Walk\textquotedblright \ }, what is the genre of the fiction? 
		\linebreak
		\myfont{A. Fantasy novel B. Fairy novel C. Mystery novel D. Historical novel} \\
		\midrule
		Role behavior & The connections between the role and his/her behavior, including the reasons for the role to do the behavior and so on. & Regarding the novel \textit{\textquotedblleft The Kite Runner\textquotedblright \ }, why did Amir win the championship in a kite competition in 1975?
		\linebreak \myfont{A. In order to get the championship prize B. To stand out in front of friends C. To win the favor of my father D. To win a bet} \\
		\midrule
		Event relation & The relations and effects of events described in a fiction, e.g., causation, correlation. & Regarding the fiction \textit{\textquotedblleft Xu Sanguan Selling Blood\textquotedblright \ }, what relationship does Xu Sanguan sell blood the second time and Yi Le injure others?
		\linebreak \myfont{A. No relationship B. Mutually exclusive relationship C. Causal relationship D. Time relationship} \\
		\midrule
		Fiction plot & The reason and background of the plot and events in a fiction.  & Regarding the novel \textit{\textquotedblleft Water Margin\textquotedblright \ }, what was the final result of Liang Shanbo's heroes fighting against the imperial court?
		\linebreak \myfont{A. Defeated and disintegrated B. Give up the fight and submit to the court. C. Win and establish a new regime D. Continue to wander around} \\
		\midrule
		Background topic & The background, era background and theme of a fiction, e.g., the values and themes conveyed by the novel. & Regarding the fiction \textit{\textquotedblleft The White Night Walk\textquotedblright \ }, what message is the novel trying to convey? 
		\linebreak
		\myfont{A. Positive energy B. Darkness of human nature C. Eternal goodness D. Money and depravity} \\
		\midrule
		Counterfactual reasoning & A situation or description that does not align with a fiction, such as a false character relationship or an event that does not exist in the fiction. & Regarding the fiction \textit{\textquotedblleft Fortress Besieged\textquotedblright \ }, what is the name of Fang Hung-chien and Tang Hsiu-fu's child? 
		\linebreak \myfont{A. Fang Hongtu B. Fang Fengyi C. Fang Feicong D. Characters do not exist} \\
        \bottomrule
	\end{tabularx}
	\caption{\label{tab: question-classify}
		Descriptions and examples of the 8 question categories. 
	}
\end{table*}

Though we have clarified the meaning of the content of each judgment node with the workers and reviewers involved in annotation and review process, mistakes still occur in the three types of questions: role behavior, event relationship and plot analysis. For example,  when we are determining whether the question is involving the occurrence of events to classify the question into role behaviour category or fiction plot category, we are referring to the conditions that the event needs to occur, such as the person and the reason. When the question is about where and when the event occurs, these are not necessary conditions but attributes of the event. Besides, when we are determining whether the question containing only event, we should first pretend answer is filled in the question and then determine if the question contains two complete events, that is, necessary conditions that the event needs to occur. 

\begin{table}[!ht]
    \centering
    \begin{tabularx}{\linewidth}{ 
			>{\raggedright\arraybackslash\hsize=1.6\hsize\linewidth=\hsize}X
			>{\centering\arraybackslash\hsize=.9\hsize\linewidth=\hsize}X
			>{\centering\arraybackslash\hsize=.7\hsize\linewidth=\hsize}X
			>{\centering\arraybackslash\hsize=.8\hsize\linewidth=\hsize}X
		}
  \toprule
         Q. Category&  Average length & Count & Ratio(\%)\\
         \midrule
         Character relationships&  27.99& 120& 19.20\\
         \midrule
         Characterization&  30.05& 184& 14.11\\
         \midrule
         Literary style&  29.80& 102& 7.82\\
         \midrule
         Role behavior&  31.80& 277& 21.24\\
         \midrule
         Event relation&  38.79& 154& 11.81\\
         \midrule
         Fiction plot&  33.89& 140& 10.74\\
         \midrule
         Background topic&  27.97& 207& 15.87\\
         \midrule
         Counterfactual reasoning&  37.07& 120& 9.20\\
         \midrule
         \textbf{Overall}&  31.97& 1304& 100\\
         \bottomrule
    \end{tabularx}
    \caption{Distribution and average length of questions in the dataset.}
    \label{tab: dataset statistic}
\end{table}

\begin{table}[!ht]
    \centering
    \begin{tabularx}{\linewidth} { 
			>{\raggedright\arraybackslash\hsize=0.8\hsize\linewidth=\hsize}X 
			>{\centering\arraybackslash\hsize=1.6\hsize\linewidth=\hsize}X 
			>{\centering\arraybackslash\hsize=0.6\hsize\linewidth=\hsize}X 
		}
        \toprule
         Category&  Subcategory& Count\\
         \midrule
         \multirow{8}{=}{Novel Type} &  Love \& Friendship& 18\\
         
         ~&  Growth \& Life& 16\\
         ~&  Society \& Human nature& 15\\
         ~& Military \& History&15\\
         ~& Crime \& Mystery&15\\
         ~& Science fiction \& Fantasy&8\\
         ~& Fables \& Philosophies&5\\
         ~& Literature \& Art&3\\
         \midrule
         \multirow{3}{=}{Character numbers}& 30k-100k&31\\
         ~& 100k-1m&58\\
         ~& over 1m&6\\
         \midrule
         \multirow{4}{=}{Publish year}& before 1900&7\\
         ~& 1900-1950&18\\
         ~& 1951-2000&43\\
         ~& 2001-now&27\\
         \bottomrule
    \end{tabularx}
    \caption{Dataset statistics.}
    \label{tab: statistic}
\end{table}

For instance, \textquotedblleft Regarding the fiction \textit{'Water Margin'}, who pulled the weeping willows\textquotedblright \ , \textquotedblleft Regarding the fiction \textit{'Water Margin'}, when did Zhishen pull the weeping willows?\textquotedblright \  and \textquotedblleft Regarding the fiction \textit{'Water Margin'}, what did Zhishen do before he pulled the weeping willows?\textquotedblright \  The first belongs to character behavior, the second is event relation, while the third belongs to fiction plot. This means that questions with similar meaning may fall into different categories when asked in different forms, and this is something to distinguish between them.

\subsection{Collecting Questions}
We design a fine-grained and strict procedure for data annotation, which ensures the quality of LFED. The procedure is well illustrated in Figure \ref{fig: workflow}. We hire eight crowdsourced workers, all senior students from the Faculty of Arts, and their reading volume can cover the 95 novels we have selected. We also hire two experts to review the answers. Both annotators and experts meet the requirements of having extensive reading experience and a strong understanding of literary works. The crowdsourced workers follow the annotation convention and our carefully defined question taxonomy (shown in Section \ref{subsec: Question Taxonomy}) to create questions and answers for their assigned fictions. Each novel are assigned to at least 2 workers. Each created data instance, including a question, four options, an answer, and associated annotations (e.g., question type according to our question taxonomy), is reviewed by two experts, and if one expert disagrees with the results, the question and feedback are sent back to the worker for recomposition until the questions can pass both experts’ reviews. 

During the expert checking process, both of them should answer a series of questions as follows:

\begin{enumerate}
    \item Are there grammatical errors or typos in the questions and answers given?
    \item If the category of the question is not counterfactual reasoning, is the category annotated by the crowdsourced worker correct according to the defined question taxonomy?
    \item If the category of the question is counterfactual reasoning, is the question contrary to the content of the novel?
    \item Is only one of the given multiple answers for the question is correct and can be selected as the final answer to the question?
\end{enumerate}

Ultimately, we observe affirmative response rates of 0.84\%, 87.92\% (1,041 out of 1,184), 98.33\% (118 out of 120) and 99.85\% to these four questions answered by reviewers, respectively. The accuracy of annotation is low relatively. This mainly stems from the fact that a novel is annotated by at least two annotators, and they have the difficulty in achieving consensus on three main question types mentioned in section \ref{subsec: Question Taxonomy}, which are role behavior, event relation and fiction plot. In addition, we also find that if a wrong question is annotated by one crowdcourced worker, the wrong question is prone to be one of the above three question types, because he/she misunderstands judgment nodes in decision-tree-style question taxonomy.

We identify questions in a format of \textquotedblleft Regarding the fiction \textit{'Fiction Name'}, which choice is wrong / right?\textquotedblright \  because these questions can not be categorized according the question taxonomy and can always be transferred to a question format adhering to the question taxonomy without changing the question meaning. Since each novel is annotated by more than one annotator, there are some redundancies in questions created by different annotators but related to the same novel. However, the proportion of such cases is very low. We filter out questions that have similar meaning and obtain 1,304 unique questions covering all 8 question categories pre-designed in our question taxonomy.

\begin{table*}[!ht]
	\centering
	\begin{tabularx}{\textwidth} { 
			>{\raggedright\arraybackslash}X
			>{\raggedright\arraybackslash}X
			>{\raggedright\arraybackslash}X
		}
		\toprule
		\textbf{Model} & \textbf{Size} &\textbf{Base Model} \\
		\midrule
		ChatGPT & 175B & instruct GPT \citep{DBLP:conf/nips/Ouyang0JAWMZASR22}\\
		ChatGLM-6B & 6.2B & GLM \citep{DBLP:conf/acl/DuQLDQY022}  \\
		BELLE-7B-0.2M & 7.1B(0.25M intructions) & bloomz-7b1-mt  \\
		BELLE-7B-2M &  7.1B(2.05M intructions) & bloomz-7b1-mt \\
		BLOOM-560M & 560M & bloomz-560m  \\
		BLOOM-1B7 & 1.7B & bloomz-1b7  \\
		\bottomrule
	\end{tabularx}
	\caption{\label{tab: models}
		Evaluated large language models.
	}
\end{table*}

\subsection{Dataset Statistics }
Table \ref{tab: dataset statistic} provides the distribution and average length of questions in the created dataset. It reveals that 277 out of 1,304 questions are on \textit{Role behavior}, accounting for 21.24\% . This is followed by questions of \textit{Chatacter relationships} (19.20\%) and \textit{Background topic} (15.87\%). 

Table \ref{tab: statistic} shows the additional statistics on LFED. First, among the 1,304 selected fictions, \textit{Love \& Friendship} emerges as the most prevalent themes, while fictions \textit{Literature \& Art} themes are relatively less common. Specifically, 66 of the fictions are originally written in Chinese, accounting for 69.47\%. Second, the majority of selected fictions (i.e., 64 fictions) are very long, containing more than 100K Chinese characters (longer than the context window of most LLMs). 6 fictions are even longer than 1M Chinese characters, far beyond the largest context window of current LLMs. Besides, Table \ref{tab: statistic} showcases the time periods of selected fictions, which cover several centuries, with the earliest dating back to the 14th century.

The average length of questions is 25.516 characters. Each question is accompanied with four answer choices. 


\begin{table*}[!ht]
	\centering
	\begin{tabularx}{\linewidth} {
			>{\raggedright\arraybackslash\hsize=1.4\hsize\linewidth=\hsize}X 
			>{\centering\arraybackslash\hsize=.9\hsize\linewidth=\hsize}X 
			>{\centering\arraybackslash\hsize=.9\hsize\linewidth=\hsize}X 
			>{\centering\arraybackslash\hsize=.9\hsize\linewidth=\hsize}X 
			>{\centering\arraybackslash\hsize=.9\hsize\linewidth=\hsize}X 
		}
		\toprule
		\multirow{2}{*}{\textbf{Model}} & \multicolumn{2}{c}{\textbf{Short Prompt}} & \multicolumn{2}{c}{\textbf{Long Prompt}} \\
		~ & zero-shot & few-shot & zero-shot & few-shot \\
		\midrule
		ChatGPT & \textbf{57.08}\% & \textbf{51.58}\% & \textbf{51.53}\% & \textbf{47.26}\% \\
		ChatGLM-6B & 39.15\% & 32.65\% & 34.14\% & 30.45\% \\
		BLOOM-560M & 31.57\% & 31.71\% & 28.46\% & 33.06\% \\
		BLOOM-1B7 & 28.05\% & 31.23\% & 29.99\% & 32.18\% \\
		BELLE-7B-0.2M & 42.18\% & 21.35\% & 38.48\% & 23.71\% \\
		BELLE-7B-2M & 39.95\% & 20.13\% & 39.78\% & 35.46\% \\
		\bottomrule
	\end{tabularx}
	\caption{\label{tab: result}
		Zero-shot results and average results under the few-shot setting (over different shots) with different prompts.
	}
\end{table*}

\section{Experiments}
We evaluated a number of LLMs on the curated dataset to investigate the capability of current LLMs on fiction comprhension. 

\subsection{Settings}
\textbf{LLMs}
We evaluated 6 large language models, which are displayed in Table \ref{tab: models}.

\begin{table*}[!ht]
	\centering
	\begin{tabularx}{\textwidth} { 
			>{\raggedright\arraybackslash\hsize=1.2\hsize\linewidth=\hsize}X 
			>{\centering\arraybackslash\hsize=.8\hsize\linewidth=\hsize}X 
			>{\centering\arraybackslash}X 
			>{\centering\arraybackslash}X
			>{\centering\arraybackslash}X 
			>{\centering\arraybackslash}X 
			>{\centering\arraybackslash}X 
			>{\centering\arraybackslash}X 
			>{\centering\arraybackslash}X 
			>{\centering\arraybackslash}X 
		}
		\toprule
		Model & { PT} & { CR}  &{ CH}  &{ LS} &{ RB}  &{ER}  & {FP}& {BT } &{CRE}   \\
		\midrule
		\multirow{2}{*}{ChatGPT} & sp & \textbf{45.00\%} & \textbf{67.39\%} & \textbf{65.69\%} & \textbf{50.54\%} & \textbf{51.30\%} & \textbf{60.71\%} & \textbf{61.84\%} & 54.17\% \\
		~ & lp & \textbf{45.83\%} & \textbf{60.87\%} & \textbf{65.69\%} & \textbf{49.46\%} & \textbf{50.65\%} & \textbf{60.00\%} & \textbf{64.73\%} & 15.00\% \\
		\multirow{2}{=}{ChatGLM-6B} & sp & 40.00\% & 46.20\% & 55.88\% & 42.96\% & 35.71\% & 36.43\% & 46.86\% & 9.17\% \\
		~ & lp & 33.33\% & 40.22\% & 38.24\% & 36.10\% & 30.52\% & 40.00\% & 46.38\% & 8.33\% \\ 
		\multirow{2}{=}{BLOOM-560M} & sp & 33.33\% & 33.15\% & 29.41\% & 28.52\% & 34.42\% & 35.71\% & 37.20\% & 20.83\% \\
		~ & lp & 31.67\% & 28.26\% & 32.35\% & 31.05\% & 21.43\% & 30.00\% & 36.23\% & 16.67\% \\
		\multirow{2}{=}{BLOOM-1B7} & sp & 28.33\% & 30.98\% & 25.49\% & 29.96\% & 33.12\% & 27.14\% & 37.68\% & 11.67\% \\
		~ & lp & 26.67\% & 34.24\% & 25.49\% & 32.13\% & 33.77\% & 30.00\%
        & 43.48\% & 14.17\% \\
		\multirow{2}{=}{BELLE-7B-0.2M} & sp & 44.17\% & 45.65\% & 46.08\% & 33.21\% & 21.43\% & 44.29\% & 43.48\% & \textbf{59.17\%} \\
		~ & lp  & 38.33\% & 38.59\% & 43.14\% & 31.77\% & 37.66\% & 37.14\% & 45.41\% & 35.83\% \\
		\multirow{2}{=}{BELLE-7B-2M} & sp & 40.00\% & 39.67\% & 47.06\% & 32.85\% & 20.78\% & 41.43\% & 44.44\% & 53.33\% \\
		~ & lp & 37.50\% & 42.39\% & 49.02\% & 33.57\% & 22.08\% & 35.00\% & 47.83\% & \textbf{50.83\%} \\
		\bottomrule
	\end{tabularx}
	\caption{\label{tab: zero-shot on different class}
		Zero-shot results over different question categories. \textbf{sp}: short prompt; \textbf{lp}: long prompt; \textbf{PT}: prompt type; \textbf{CR}: Characterization; \textbf{CH}: Characterization; \textbf{LS}: Literary style; \textbf{RB}: Role behavior; \textbf{ER}: Event Relation; \textbf{FP}: Fiction plot; \textbf{BT}: Background topic; \textbf{CRE} Counterfactual reasoning.
	}
\end{table*}

\textbf{Prompts}
We conducted zero-shot and few-shot tests of LLMs on the LFED. For zero-shot tests, we utilized two types of prompts: long prompt and short prompt. The short prompt is in a uniform format of \textquotedblleft Select the desired answer based on the given question\textquotedblright \  while the long prompt is \textquotedblleft Give you a multiple-choice question about a fiction, and you need to provide an answer. The provided string can be divided into three parts: the first part represents the title of the fiction, the second part is a question about the fiction, the third part includes four answer choices. Please only output the answer indicator (e.g., A, B, C or D).\textquotedblright \  In the few-shot tests, we augmented the long and short prompts used in the zero-shot tests by providing $n$ examples based on the number of shots, where $n$ is an integer from 0 to 5. And we make sure that the examples provided to LLMs in the prompt are from different novels.

The input for all large language models consisted of the prompt, question, answer choices, and the suffix \textquotedblleft the correct answer is:\textquotedblright \ .

\textbf{Evaluation Process} Given the prompt, LLMs may not only output the answer choice indicator. For example, we find that ChatGPT under the zero-shot setting usually output not only the answer indicator but also rationals for the answer. We hence provide a script to deal with this issue. We run this process at most three times, or we will treat as LLMs answer wrong, because some model outputs will stay the same after multiple iterations: 
\begin{enumerate}
\item  Checking whether the output contains only one answer indicator. If so, the corresponding answer indicator is treated as the answer.
\item  If the output contains multiple answer indicators, we choose the indicator occurring most frequently in the output as the answer.
\item If the above conditions are not met, we change the suffix to \textquotedblleft According to the above question, please output the answer directly and do not output any rationals. \textquotedblright \ .
\end{enumerate}

All text inputs to LLMs are in Chinese, as we are evaluating Chinese LLMs.

\textbf{Evaluation Integrity} The potential exposure of novels in LFED to LLMs during their pre-training stage is acknowledged, but measures have been taken to ensure that the evaluation reflects the capabilities of LLMs in \textquotedblleft truly understanding\textquotedblright \  these novels. These measures include:
\begin{enumerate}
    \item Manual Annotation and Review: This guarantees that the created questions are unique and not present in  pre-training and alignment training data of LLMs. Additionally, we require human annotators to focus on various cognitive abilities when producing questions. 
    \item Diverse Question Design: We encourage human annotators to create diverse questions, including those on assessing abilities in fact understanding, logical reasoning, etc. 
\end{enumerate}

\begin{table*}[!ht]
	\centering
	\begin{tabularx}{\textwidth} { 
			>{\raggedright\arraybackslash\hsize=1.2\hsize\linewidth=\hsize}X 
			>{\centering\arraybackslash\hsize=.8\hsize\linewidth=\hsize}X 
			>{\centering\arraybackslash}X 
			>{\centering\arraybackslash}X
			>{\centering\arraybackslash}X 
			>{\centering\arraybackslash}X 
			>{\centering\arraybackslash}X 
			>{\centering\arraybackslash}X 
			>{\centering\arraybackslash}X 
			>{\centering\arraybackslash}X 
		}
		\toprule
		Model & { PT} & { CR}  &{ CH}  &{ LS} &{ RB}  &{ER}  & {FP}& {BT } &{CRE}   \\
		\midrule
		\multirow{2}{*}{ChatGPT} & sp & \textbf{44.27\%} & \textbf{61.85\%} & \textbf{68.63\%} & \textbf{45.63\%} & \textbf{48.44\%} & \textbf{55.14\%} & \textbf{59.71\%} & \textbf{29.00\%} \\
		~ & lp & \textbf{40.35\%} & \textbf{54.89\%} & \textbf{61.76\%} & \textbf{47.00\%} & \textbf{47.01\%} & \textbf{53.00\%} & \textbf{53.72\%} & \textbf{20.33\%} \\
		\multirow{2}{=}{ChatGLM-6B} & sp & 35.04\% & 40.54\% & 37.84\% & 31.99\% & 34.42\% & 35.29\% & 37.10\% & 9.00\% \\
		~ & lp & 33.33\% & 36.74\% & 31.57\% & 31.12\% & 31.69\% & 32.43\% & 37.87\% & 8.83\% \\ 
		\multirow{2}{=}{BLOOM-560M} & sp & 36.08\% & 34.46\% & 33.14\% & 34.30\% & 34.68\% & 36.43\% & 39.61\% & 5.00\% \\
		~ & lp & 34.88\% & 35.00\% & 33.33\% & 35.96\% & 34.81\% & 40.29\% & 43.19\% & 7.00\% \\  
		\multirow{2}{=}{BLOOM-1B7} & sp & 33.17\% & 36.96\% & 35.88\% & 33.57\% & 37.14\% & 32.14\% & 37.10\% & 3.83\% \\
		~ & lp & 33.35\% & 37.07\% & 32.94\% & 34.08\% & 37.14\% & 36.00\% & 40.68\% & 6.17\% \\  
		\multirow{2}{=}{BELLE-7B-0.2M} & sp & 23.07\% & 24.46\% & 24.51\% & 19.78\% & 19.61\% & 19.29\% & 17.29\% & 22.83\% \\
		~ & lp & 21.90\% & 25.65\% & 29.41\% & 25.34\% & 27.01\% & 24.43\% & 20.77\% & 15.17\% \\  
		\multirow{2}{=}{BELLE-7B-2M} & sp & 21.54\% & 20.11\% & 23.14\% & 18.84\% & 17.40\% & 17.00\% & 14.98\% & 28.00\% \\
		~ & lp & 41.05\% & 39.89\% & 44.31\% & 34.08\% & 37.79\% & 33.86\% & 44.35\% & 8.33\% \\
		\bottomrule
	\end{tabularx}
	\caption{\label{tab: few-shot on different class}
		Average few-shot results over different question categories. \textbf{sp}: short prompt; \textbf{lp}: long prompt; \textbf{PT}: prompt type; \textbf{CR}: Characterization; \textbf{CH}: Characterization; \textbf{LS}: Literary style; \textbf{RB}: Role behavior; \textbf{ER}: Event Relation; \textbf{FP}: Fiction plot; \textbf{BT}: Background topic; \textbf{CRE} Counterfactual reasoning.
	}
\end{table*}

\subsection{Results}
\paragraph{Overall} The dataset is mainly developed to evaluate two capabilities of LLMs: the ability of applying knowledge of long novels obtained during training and the ability of reasoning over long novels. The zero- and few-shot results do have some implications for this evaluation goal. Besides, the design of long and short prompts can also give some clues in evaluating LLMs. Results are presented in Table \ref{tab: result}. We observe that ChatGPT achieves the highest performance under the zero- and few-shot setting with both short and long prompts, reaching accuracy of 57.08\% in zero-shot with short prompt. While the accuracy gaps between other evaluated LLMs and CharGPT are around 20\% or even greater.

Zero-shot results over different question categories are reported in Table \ref{tab: zero-shot on different class}. We observe that ChatGPT has the best performance in other categories except counterfactual reasoning. On the counterfactual reasoning, Belle-7b-0.2M performed best with short prompt, reaching accuracy of 59.17\%, while Belle-7b-2M performed best with long prompt, reaching accuracy of 50.83\%.

Table \ref{tab: few-shot on different class} displays average few-shot results of the evaluated LLMs over different categories. Surprisingly, ChatGPT achieves the highest accuracy across all question categories with short prompts and long prompts. 


\paragraph{Further Analysis} As shown in Table \ref{tab: result}, among evaluated LLMs, BLOOM series models may have deeper understand of long prompts as these models has higher performance in few-shot setting or with long prompt, while other evaluated models perform poorly in these settings.

\begin{figure*}[!ht]
	\centering
	\includegraphics[width = \linewidth]{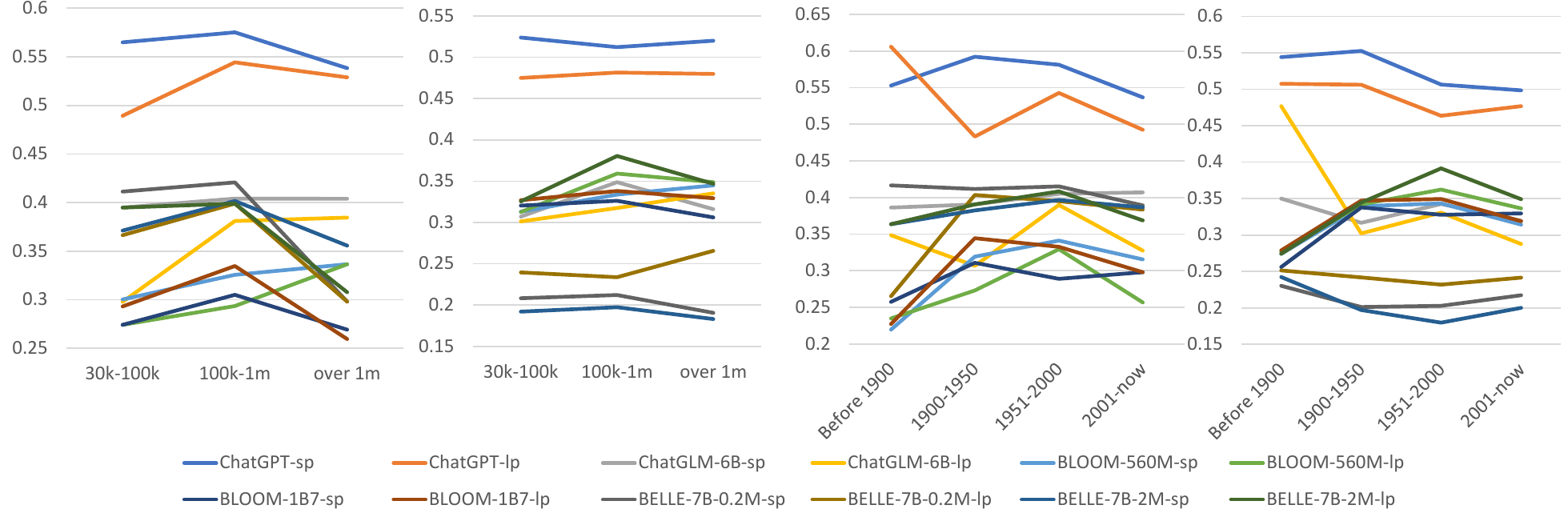}
	\caption{\label{fig: siheyi}Results on different novel attributions under the zero- and few-shot setting. The suffixes \textbf{-sp} and \textbf{-lp} in the model name represent short prompt and long prompt respectively. The left two subfigures demonstrate results on different range of chatacter numbers under the zero- and few-shot setting respectively. The right two subfigures demonstrate results on different range of publish years under the zero- and few-shot setting respectively.}
\end{figure*}

\begin{figure}
    \centering
    \includegraphics[width=\linewidth]{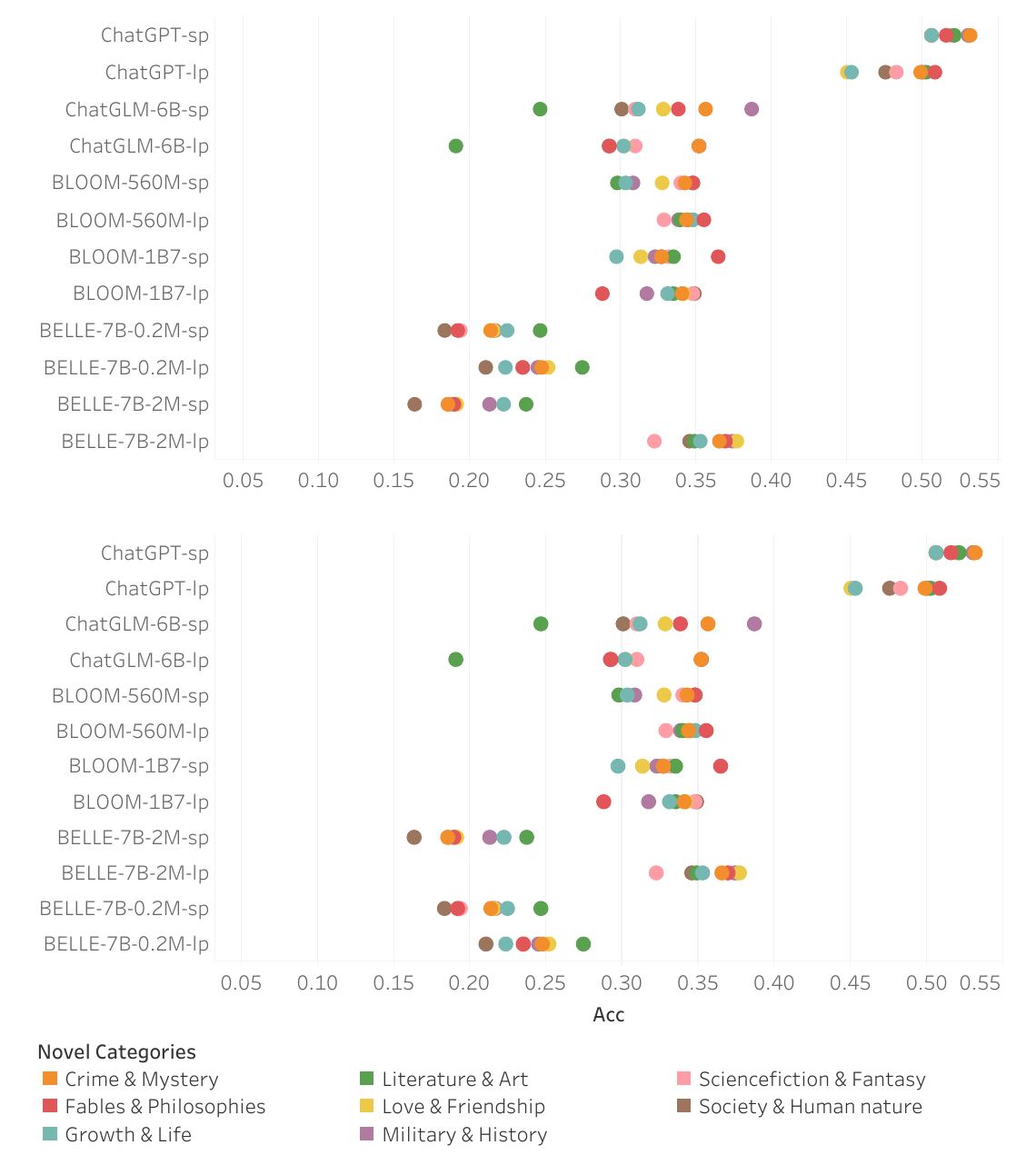}
    \caption{Results on different novel types under the zero- and few-shot setting. The suffixes \textbf{-sp} and \textbf{-lp} in the model name represent short prompt and long prompt respectively. The top figure shows zero-shot results while the bottom one demonstrates few-shot results.}
    \label{fig: res_differentFT}
\end{figure}

Let's compare Table \ref{tab: zero-shot on different class} and Table \ref{tab: few-shot on different class} together. We can see that ChatGPT has the best overall performance among the evaluated LLMs across all experiments, and leads almost all other evaluated LLMS by more than 10\% in most experiments. The lowest results on counterfactual reasoning and event relation indicate that LLMs struggle on reasoning on extremely long documents while relatively good results on characterization, background topic and literature style suggest that LLMs performs better on knowledge acquisition on novels than reasoning on novels. Surprisingly, BELLE-7Bt-0.2M and BELLE-7B-2M achieve high accuracies on counterfactual reasoning. This may be because a large proportion of questions in the training data which are \textquotedblleft Unanswerable\textquotedblright \ , rather than having strong reasoning capabilities because the two models perform generally in event relation, a category of questions that require reasoning. This makes these two models more inclined to choose \textquotedblleft unanswerable\textquotedblright \  as the answer, so the accuarcy on counterfactual reasoning are relatively high.

Besides, results shown in  Table \ref{tab: zero-shot on different class} and Table \ref{tab: few-shot on different class} indicate significant variability in performance across different question types and prompt lengths. For instance, it is noteworthy that the short prompts tend to yield higher accuracy in certain categories like Characterization  and Fiction Plot, whereas long prompts seem to facilitate better performance in Background Topic  and Counterfactual Reasoning.

We studied whether the number of characters of selected novel and the year of publication the novel was published would have an impact on the results. The results are shown in Figure \ref{fig: siheyi}. Most models perform better on novels with character numbers between 100,000 and 1 million in zero- and few-shot setting. It's possible that it's more difficult to acquire knowledge and make inferences in shorter or longer novels. Because shorter novels may have omissions in the storyline, this will make it difficult for LLMs to acquire knowledge. Extreme long novels (over 1 million characters in our dataset) have complex character relationships and storylines, which can make it difficult for LLMs to extract the right knowledge and various relations. Through the information obtain from the right two subfigures in Figure \ref{fig: siheyi}, we can find that there is no obvious trend in the results of the model on novels of different publication years, which shows that the differences in language habits reflected in different times of novels do not affect the evaluation of the model.

We also evaluated performance by the type of novels. Results (see Figure \ref{fig: res_differentFT}) show that the performance of ChatGLM-6B varies greatly across different novel types with both long and short prompts, and it performs poorly on literary style questions under both zero- and few-shot setting. Besides, we can also observe that the performance of BELLE-7B-0.2M on different novel categories decreases substantially with long prompt when several demonstrations are provided, indicating that BELLE-7B-0.2M does not understand long prompt and learn from demonstrations well.

\section{Conclusion}
We have presented LFED, a literary fiction evaluation dataset which consisting of 1,304 questions from 95 fictions, designed to evaluate the reasoning capability of large language models. Our dataset curation features carefully selected fictions, a question taxonomy aiming at diversity and coverage, engaged workers and reviewers. Experiments demonstrate that the dataset is challenging for state-of-the-art LLMs under both zero- and few-shot setting.

\section*{Acknowledgements}
The present research was partially supported by Huawei and Zhejiang Lab (No. 2022KH0AB01). We would like to thank the anonymous reviewers for their insightful comments.

\section{Bibliographical References}\label{sec:reference}

\bibliographystyle{lrec-coling2024-natbib}
\bibliography{lrec-coling2024-example}


\end{document}